# Sentiment Analysis in the News

Alexandra Balahur[1], Ralf Steinberger[2], Mijail Kabadjov[2], Vanni Zavarella[2],

Erik van der Goot[2], Matina Halkia[2], Bruno Pouliquen[2], Jenya Belyaeva[2]

[1] University of Alicante, Department of Software and Computing Systems
Ap. de Correos 99, E-03080 Alicante, Spain
[2] European Commission – Joint Research Centre
IPSC - GlobeSec - OPTIMA (OPensource Text Information Mining and Analysis)
T.P. 267, Via Fermi 2749
21027 Ispra (VA), Italy

abalahur@dlsi.ua.es,
{Ralf.Steinberger, Mijail.Kabadjov, Erik.van-der-Goot, Matina.Halkia,Bruno.Pouliquen}@jrc.ec.europa.eu,
{Vanni.Zavarella, Jenya.Belyaeva}@ext.jrc.ec.europa.eu

## Abstract

Recent years have brought a significant growth in the volume of research in sentiment analysis, mostly on highly subjective text types (movie or product reviews). The main difference these texts have with news articles is that their target is clearly defined and unique across the text. Following different annotation efforts and the analysis of the issues encountered, we realised that news opinion mining is different from that of other text types. We identified three subtasks that need to be addressed: definition of the target; separation of the good and bad news content from the good and bad sentiment expressed on the target; and analysis of clearly marked opinion that is expressed explicitly, not needing interpretation or the use of world knowledge. Furthermore, we distinguish three different possible views on newspaper articles – author, reader and text, which have to be addressed differently at the time of analysing sentiment. Given these definitions, we present work on mining opinions about entities in English language news, in which (a) we test the relative suitability of various sentiment dictionaries and (b) we attempt to separate positive or negative opinion from good or bad news. In the experiments described here, we tested whether or not subject domain-defining vocabulary should be ignored. Results showed that this idea is more appropriate in the context of news opinion mining and that the approaches taking this into consideration produce a better performance.

## 1. Introduction

Most work on opinion mining has been carried out on subjective text types such as blogs and product reviews. Authors of such text types typically express their opinion freely. The situation is different in news articles: many newspapers (with the exception of a few tabloids that are monitored by EMM) at least want to give an impression of objectivity so that journalists will often refrain from using clearly positive or negative vocabulary. They may resort to other means to express their opinion, such as embedding statements in a more complex discourse or argument structure, they may omit some facts and highlight others, they may quote other persons who say what they feel, etc. Automatically identifying sentiment that is not expressed lexically is rather difficult, but lexically expressed opinion can be found in news texts, even if it is less frequent than in product or film reviews. Another difference between reviews and news is that reviews frequently are about a relatively concrete object (referred to as the 'target'), while news articles may span larger subject domains, more complex event descriptions and a whole range of targets (e.g. various, even opposing, politicians). Unpublished in-house experiments on document-level sentiment analysis (counting stronger and weaker positive and negative words in the whole article) led us to believe that it is very important to clearly identify the target of any sentiment expressed and to restrict the analysis to the immediate context of the target (Balahur and Steinberger, 2009). We have also observed that automatic opinion mining systems usually identify negative opinion values about entities when these were mentioned in the context of negative news, such as, for instance, the outbreak of the world financial crisis in 2008. This negative spike is mostly independent of the role of an entity in the events, i.e. the sentiment value towards a person may be negative even if this person is attempting to act positively in the event. For these reasons, we have focused in our recent opinion mining experiments, presented here, on considering smaller and larger word windows around entities, and we have attempted to separate positive and negative sentiment from good and bad news.

## 2. The EMM News Data

The EMM applications NewsBrief and MedISys categorise the news into one or more of several hundred subject domain classes, including, for instance, natural disasters, security, finance, nuclear issues, various diseases, organisations, countries, regions, specific conflicts, etc. Categorisation is achieved by (often user-defined) Boolean search word expressions or by using lists of search words with varying (positive or negative) weights and a threshold (Steinberger et al. 2009). These category-defining word lists will thus contain terms such as 'disaster', 'tsunami' and 'crisis', etc., which are likely to also be found in lists of sentiment vocabulary. The idea we followed up in our experiments is to exclude those category-defining words from our sentiment analysis that are part of the category definitions of the subject domains with which the news article was tagged. The category definitions may not contain all content words that are also sentiment vocabulary and a more complete hand-produced list might be more efficient. However, the advantage of using the existing category definitions is that they are all ready-made for dozens of languages, making it simple to use the same method for sentiment analysis in many more languages without much effort, should the approach be successful.

From the news in 13 languages, an average 3165 reported speech quotations per day are automatically extracted (Pouliquen et al., 2007). The person issuing the quotation is extracted, and so is any entity that is being mentioned inside the quotation. In the experiments presented here, we test our methods on these automatically extracted quotations, although nothing would stop us from applying them to any other text segment. The reason for using quotations is that the text in quotes is usually more subjective than the other parts of news articles. We also know for quotes who the person is that made the statement (referred to as the source of the opinion statement) and – if the speaker makes reference to another entity within the quotation – we have a clue about the possible target (or object) of the sentiment statement.

Although at this point we only employ the presented algorithm on quotes, the main objective of our research is to determine the best approach to detecting sentiment in the news in general. Such an algorithm can subsequently be employed in all news texts, not only quotes.

## 3. Related Work

Subjectivity analysis is defined by Wiebe (1994) as the "linguistic expression of somebody's opinions, sentiments, emotions, evaluations, beliefs and speculations". In her definition, the author was inspired by the work of the linguist Ann Banfield (Banfield, 1982), who defines as subjective the "sentences that take a character's point of view (Uspensky, 1973)" and that present private states (Quirk, 1985) (i.e. states that are not open to objective observation or verification) of an experiencer, holding an attitude, optionally towards an object.

(Esuli and Sebastiani, 2006) define opinion mining as a recent discipline at the crossroads of information retrieval and computational linguistics which is concerned not with the topic a document is about, but with the opinion it expresses.

(Dave et al., 2003), define an opinion mining system as one that is able to "process a set of search results for a given item, generating a list of product attributes (quality, features, etc.) and aggregating opinions about each of them (poor, mixed, good)." Opinion mining, in this context, aims therefore at extracting and analysing judgements on various aspects of given products. A similar paradigm is given by (Hu and Liu, 2004), which the authors entitle feature-based opinion mining.

(Kim and Hovy, 2005) define opinion as a quadruple (Topic, Holder, Claim, Sentiment), in which the Holder believes a Claim about the Topic, and in many cases associates a Sentiment, such as good or bad, with the belief. The authors distinguish among opinions with sentiment and opinions without sentiment and between directly and indirectly expressed opinions with sentiment. In other approaches, capturing favourability versus unfavourability, support versus opposition, criticism versus appreciation, liking versus disliking, even bad versus good news classification were considered to be sentiment analysis. However, at the moment of annotating sentiment in newspaper articles, we have seen that combining all these aspects together did not help to clarify what the task was and how annotation should be done. Even in the case of quotes, which are short pieces of text where the source was known and the possible targets were identified, expressions of opinion that needed some kind of interpretation or knowledge of the situation fell short of agreement, due to personal convictions, background and so on.

## 4. Experiments and evaluation

### 4.1 Redefining the task

To clarify the task of opinion mining from news, we selected a collection of 1592 quotes (reported speech) from newspaper articles in English, whose source and target were known (their extraction patterns are designed with that scope) which we set out to annotate. A histogram of the quotes' length is shown in Figure 1.

The first experiments had an inter-annotator agreement of under 50%. Specifying that just the sentiment on the target should be annotated and separated from the good and bad news that was described led to an increase in the agreement up to 60%. We realised that by delimiting a few aspects, the task became much clearer. These aspects included not using one's background knowledge or interpreting what is said. The original data set we decided to annotate contained 1592 quotes extracted from news in April 2008. The average final agreement was 81%, between 3 pairs of two annotators each.

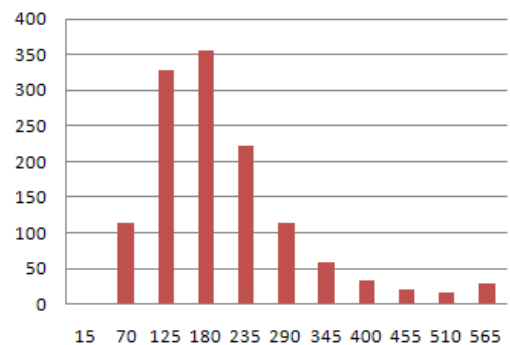

Figure 1: Histogram of the quotes' length

|  | No. quotes | No. agr. quotes | No. agr. neg. quotes | No. agr. pos. quotes | No. agr. obj. quotes |
|---|---|---|---|---|---|
|  | 1592 | 1292 | 234 | 193 | 865 |
| Agr (%) |  | 81% | 78% | 78% | 83% |

Table 1: Results of the data annotation

The result of the annotation guidelines and labelling process was a corpus in which we agreed what sentiment was and what it was not. The number of agreed sentiment-containing quotes was one third of the total number of agreed quotes, showing that only clear, expressly stated opinion was marked, i.e. opinions that required no subjective interpretation from the annotator's part. The result of our labelling showed that in the case of newspapers, it is mandatory to distinguish between three different "components": the *author*, the *reader* and the *text* itself (Figure 2).

While *the author* might convey certain opinions, by omitting or stressing upon some aspect of the text and by thus inserting their own opinion towards the facts, the spotting of such phenomena is outside the aim of sentiment analysis as we have defined it. Instead, such phenomena should be analysed as part of work on perspective determination or news bias research. From the *reader's point of view*, the interpretations of the text can be multiple and they depend on the personal background knowledge, culture, social class, religion etc. as far as what is normal (expected) and what is not are concerned. Lastly, the opinion stated *strictly in the text* is the one that one should concentrate on at this level, being expressed directly or indirectly, by the source, towards the target, with all the information needed to draw this conclusion on polarity present in the text.

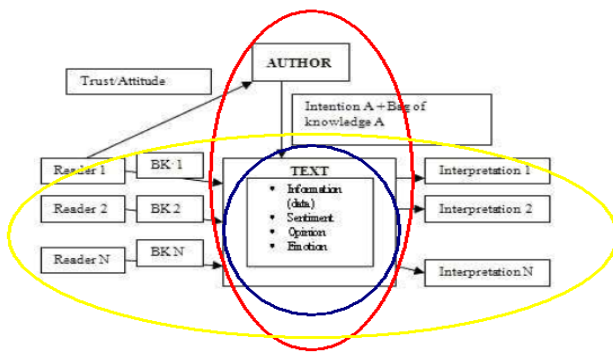

Figure 2: The three components of text opinion

From the author and the reader's perspective and not from the text's pure informational point of view, opinion is conveyed through facts that are interpretable by the emotion they convey. However, emotions are not universal in their meaning. They are determined socially, culturally and historically. There are general emotions, but most of the times they relate to the norms, their significance and the cultural environment. Emotions imply an evaluation, which is both cognitive and affective, of a behaviour, with respect to a norm and the mutual expectation it raises. Some norms are common sense and are accepted and understood by all. Normative expectations link the behaviour (reaction) to a meaning and on this ground, by the understanding it is given. From the reader's point of view, sentiment analysis would be defined as *the assessment of a "target", based on its characteristics and factual information related to it, according to whether or not the results of the assessments are "according to" or "against" the "norm"* (their personal understanding and approval of what is "good" and "bad" in a certain situation).

From the author's point of view, news bias or perspective determination should be concerned with discovering the ways in which expression of facts, word choice, omissions, debate limitations, story framing, selection and use of sources of quotes and the quote boundaries, for example, conveys a certain sentiment or not. The sentiment content of the text, finally, is what is expressly stated, and not what is left to be understood between the lines. Our effort focuses on detecting this last aspect.

**4.2 Experiments**

In order to measure the impact of our defined task, we performed different experiments on the set of 1292 quotes on which agreement has been reached. Out of these 1292, the target was successfully identified by the sentiment analysis system in 1114 quotes (direct mentions of the *target* through the name or its title). The baseline we compare against is the percentage of quotes pertaining to the largest class of quotes – objective, which represents 61% of our corpus.

According to the approach we settled on, we wanted to make sure that: a) we estimate the opinion on the target of the quote (by computing the opinion in windows of words between the mentions of the entity), b) we eliminate the bad versus good news content (by eliminating those words which are both sentiment-bearing words and words that are part of EMM category definitions, from now on called *category words*). Given that we are faced with the task of classifying opinion in a general context, we employed a simple, yet efficient approach, presented in (Balahur et al., 2009). At the present moment, there are different lexicons for affect detection and opinion mining. In order to have a more extensive database of affect-related terms, in the following experiments we used *WordNet Affect* (Strapparava and Valitutti, 2004), *SentiWordNet* (Esuli and Sebastiani, 2006), *MicroWNOp* (Cerini et al, 2007). Additionally, we used an in-house built resource of opinion words with associated polarity, which we denote by *JRC Tonality*. Each of the employed resources was mapped to four categories, which were given different scores: positive (1), negative (-1), high positive (4) and high negative (-4). The score of each of the quotes was computed as sum of the values of the words identified around the mentions of the entity that was the target of the quote, either directly (using the name), or by its title (e.g. Gordon Brown can be referred to as "*Gordon*", as "*Brown*" or as "*the British prime-minister*") [1] . The experiments conducted used different windows around the mentions of the target, by computing a score of the opinion words identified and eliminating the words that were at the same time opinion words and category words (e.g. *crisis*, *disaster*).

---

[1] For the full details on how the names and corresponding titles are obtained, please see (Pouliquen and Steinberger, 2009).

Table 2 presents an overview of the results obtained using different window sizes and eliminating or not the category words in terms of *accuracy* (number of quotes that the system correctly classified as positive, negative or neutral, divided by the total number of quotes). As it can be seen, the different lexicons available performed dramatically different and the impact of eliminating the alert words was significant for some resources or none for others, i.e. in those cases where there were no category words that coincided with words in the respective lexicon.

| Word window | W or W/O Alerts | JRC Tonality | Micro WN | WN Affect | Senti WN |
|---|---|---|---|---|---|
| Whole text | W Alerts | 0.47 | 0.54 | 0.21 | 0.25 |
|  | W/O Alerts | 0.44 | 0.53 | 0.2 | 0.2 |
| 3 | W Alerts | 0.51 | 0.53 | 0.24 | 0.25 |
|  | W/O Alerts | 0.5 | 0.5 | 0.23 | 0.23 |
| 6 | W Alerts | 0.63 | 0.65 | 0.2 | 0.23 |
|  | W/O Alerts | 0.58 | 0.6 | 0.18 | 0.15 |
| 6 | W Alerts | 0.82 | | 0.2 | 0.23 |
|  | W/O Alerts | 0.79 | | 0.18 | 0.15 |
| 10 | W Alerts | 0.61 | 0.64 | 0.22 | 0.2 |
|  | W/O Alerts | 0.56 | 0.64 | 0.15 | 0.11 |

Table 2: Accuracy obtained using different lexicons, window sizes and alerts

As we can see from the difference in the results between the opinion mining process applied to the whole text and applied only to text spans around named entities, computing sentiment around the mentions of the entity in smaller window sizes performs better than computing the overall sentiment of texts where the entities are mentioned. From our experiments, we could notice that some resources have a tendency to over-classify quotes as negative (WordNet Affect) and some have the tendency to over-classify quotes as positive (SentiWordNet). We have performed evaluations using combinations of these four lexicons. The best results we obtained were using the combination of JRC Tonality and MicroWN, on a window of 6 words; in this case, the accuracy we obtained was 82%. As we can see, the majority of the resources used did not pass the baseline (61%), which shows that large lexicons do not necessarily mean an increase in the performance of systems using them.

### 4.3 Error analysis

Subsequently to the evaluation, we have performed an analysis of the cases where the system fails in correctly classifying the sentiment of the phrase or incorrectly classifying it as neutral. The largest percentage of failures is represented by quotes which are erroneously classified as neutral, because no sentiment words are present to account for the opinion in an explicit manner (e.g. "We have given X enough time", "He was the one behind all these atomic policies", "These revelations provide, at the very least, evidence that X has been doing favours for friends", "We have video evidence that activists of the X are giving out food products to voters") or the use of idiomatic expressions to express sentiment (e.g. "They have stirred the hornet's nest"). Errors in misclassifying sentences as positive instead of negative or vice-versa were given by the use of irony (e.g. "X seemed to offer a lot of warm words, but very few plans to fight the recession"). Finally, quotes were misclassified as positive or negative (when they should in fact be neutral) because of the presence of a different opinion target in the context (e.g. "I've had two excellent meetings with X", "At the moment, Americans seem willing to support Y in his effort to win the war", "everyone who wants Y to fail is an idiot, because it means we're all in trouble", "The chances of this strategy announced by X are far better than the purely military strategy of the past...") or the use of anaphoric references to the real target.

All these problems require the implementation of specific methods to tackle them. Thus, firstly, the opinion lexicons should be extended to contain concepts which implicitly imply an assessment of the target because they are concepts we employ in our everyday lives (e.g. "hunger, food, approval"). Secondly, expressions that are frequently used in a language to describe "good" and "bad" situations have to be added to the opinion lexicon (e.g. "stir the hornet's nest", "take the bull by the horns"). Irony is difficult to detect in text; however, when dealing with a larger context, the polarity of such pieces of text could be determined in relation to that of the surrounding sentences. Further on, we are researching on methods to determine the target of the opinion using Semantic Roles; thus, the judgement on the opinion expressed can be improved. Finally, resolving co-reference using a standard tool should in theory lead to a higher performance of the opinion mining system. However, in practice, from our preliminary experiments, the performance of the opinion mining system decreases when employing anaphora resolution tool.

### 5. Conclusions and future work

In this paper, we summarised our insights regarding sentiment classification for news and applied different methods to test the appropriateness of different resources and approaches to the task defined. We have seen that there is a need to clearly define, before the annotation is done, what the source and the target of the sentiment are, subsequently separate the good and bad *news content* from the good and bad *sentiment* expressed on the target and, finally, annotate only clearly marked opinion that is expressed explicitly, not needing interpretation or the use of world knowledge. We have furthermore seen that there are three different possible views on newspaper articles – author, reader and text – and they have to be addressed differently at the time of analysing sentiment. We have performed experiments in this direction, by using categories to separate good and bad news content from the opinionated parts of the text. We also evaluated our approach using different lexicons in diverse combinations, and word windows.

We have shown that this simple approach produces good results when the task is clearly defined. Future work includes evaluating the impact of using negation and valence shifters and the use of other methods that have been proven efficient, such as machine learning using similarity with annotated corpora (Balahur et al, 2009) or syntactic patterns (Riloff and Wiebe, 2003). We also plan to extend the lexica used with different concepts that are intrinsically referring to a positive or negative situation and include target detection. Last, but not least, we are assessing methods to extend the lexicons for additional languages and subsequently compare opinion trends across sources and time.

## References

Balahur, A., Steinberger, R., Van der Goot, E., Pouliquen, B., Kabadjov, M. (2009). Opinion Mining on Newspaper Quotations. *Proceedings of the workshop 'Intelligent Analysis and Processing of Web News Content' (IAPWNC),* held at the 2009 IEEE/WIC/ACM International Conferences on Web Intelligence and Intelligent Agent Technology. Milano, Italy, 2009.

Balahur, A., Steinberger, R. (2009). Rethinking Opinion Mining in News: from Theory to Practice and Back. In *Proceedings of the 1st Workshop on Opinion Mining and Sentiment Analysis, Satellite to CAEPIA 2009*.

Banfield, A. (1982). *Unspeakable sentences: Narration and Representation in the Language of Fiction.* Routledge and Kegan Paul.

Belyaeva, E., Van Der Goot, E. (2009). News bias of online headlines across languages. The study of conflict between Russia and Georgia. August 2008. *Rhetorics of the Media. Conference Proceedings* Lodz University Publishing House.

Cerini, S. , V. Compagnoni, A. Demontis, M. Formentelli and G. Gandini. (2007). *Language resources and linguistic theory: Typology, second language acquisition, English linguistics*, chapter Micro-WNOp: A gold standard for the evaluation of automatically compiled lexical resources for opinion mining. Franco Angeli Editore, Milano, IT. 2007.

Dave, K., Lawrence, S., Pennock, D.M. (2003). Mining the peanut gallery: Opinion extraction and semantic classification of product reviews. In P*roceedings of WWW*, pp. 519–528, 2003.

Esuli, A. and F. Sebastiani. (2006). SentiWordNet: A Publicly Available Resource for Opinion Mining. In *Proceedings of the 6th International Conference on Language Resources and Evaluation, LREC 2006*, Italy. 2006.

Esuli, A. and Sebastiani, F. (2006). SentiWordNet: A publicly available resource for opinion mining. In *Proceedings of the 6th International Conference on Language Resources and Evaluation*

Fortuna Blaž, Carolina Galleguillos and Nello Cristianini. (2009). *Detecting the bias in media with statistical learning methods Text Mining: Theory and Applications,* Taylor and Francis Publisher, 2009.

Goleman, D. (1995). *Emotional Intelligence.* Bantam Books.

Kim, S.-M. and Hovy, E. (2004). Determining the Sentiment of Opinions. In Proceedings of COLING 2004.

Pouliquen Bruno, Ralf Steinberger & Clive Best (2007). Automatic Detection of Quotations in Multilingual News. In *Proceedings of the International Conference Recent Advances in Natural Language Processing (RANLP'2007)*, pp. 487-492. Borovets, Bulgaria, 27-29.09.2007.

Pouliquen, B and Steinberger, R. (2009). Automatic Construction of Multilingual Name Dictionaries. In Cyril Goutte, Nicola Cancedda, Marc Dymetman & George Foster (eds.): *Learning Machine Translation.* pp. 59-78. MIT Press - Advances in Neural Information Processing Systems Series (NIPS).

Quirk, R. (1985). *A Comprehensive Grammar of the English Language.* Longman Publishing House.

Ratner, C. (2000). A cultural-psychological analysis of emotions. *Culture and Psychology*, (6).

Riloff, E. and Wiebe, J. (2003). Learning Extraction Patterns for Subjective Expressions. In *Proceedings of the 2003 Conference on Empirical Methods in Natural Language Processing (EMNLP-03)*.

Steinberger, R., Pouliquen, B., Van der Goot, E. (2009). An Introduction to the Europe Media Monitor Family of Applications. In: Fredric Gey, Noriko Kando & Jussi Karlgren (eds.): *Information Access in a Multilingual World - Proceedings of the SIGIR 2009 Workshop (SIGIR-CLIR'2009)*, pp. 1-8. Boston, USA. 23 July 2009.

Strapparava, C. and Mihalcea, R. (2007). Semeval 2007 task 14: Affective text. In *Proceedings of ACL 2007*.

Strapparava, C. and Valitutti, A. (2004) WordNet-Affect: an affective extension of WordNet. In *Proceedings of the 4th International Conference on Language Resources and Evaluation (LREC 2004)*, Lisbon, May 2004, pp. 1083-1086. 2004.

Uspensky, B. (1973). A Poetics of Composition. University of California Press, Berkeley, California.

Wiberg, M. (2004). *The Interaction Society: Theories Practice and Supportive Technologies*. Idea Group Inc.

Wiebe, J. (1994). Tracking point of view in narrative. Computational Linguistics, 20.